\documentclass[preprint,12pt]{elsarticle}

\usepackage{amssymb}
\usepackage{amsmath}
\usepackage[hidelinks]{hyperref}

\usepackage{xcolor}

\journal{Information Fusion}

\begin{document}

\begin{frontmatter}



\title{UD-SfPNet: An Underwater Descattering Shape-from-Polarization Network for 3D Normal Reconstruction}


%\author{Puyun Wang, Kaimin Yu,Huayang He, Feng Huang, Xianyu Wu, Yating Chen} %% Author name
\author[FZU]{Puyun Wang}
\author[FZU]{Kaimin Yu}
\author[MoT]{Huayang He}
\author[FZU]{Feng Huang}
\author[FZU]{Xianyu Wu\corref{cor1}}
\ead{xwu@fzu.edu.cn}
\author[Tsinghua]{Yating Chen\corref{cor1}}
\ead{chenyating@mail.tsinghua.edu.cn}

\cortext[cor1]{Corresponding authors.}

\affiliation[FZU]{organization={School of Mechanical Engineering and Automation, Fuzhou University},%Department and Organization
            city={Fuzhou},
            postcode={35018}, 
            state={Fujian},
            country={China}}
	
\affiliation[MoT]{organization={The Research Institute of Highway, Ministry of Transport},%Department and Organization
%	addressline={}, 
	city={Beijing},
	postcode={100088}, 
%	state={Beijing},
	country={China}}
	
\affiliation[Tsinghua]{organization={The State Key Laboratory of Precision Measurement Technology and Instruments, Department of Precision Instruments, Tsinghua University},%Department and Organization
	%	addressline={}, 
	city={Beijing},
	postcode={100084}, 
%	state={Fujian},
	country={China}}

%% Abstract
\begin{abstract}

Underwater optical imaging is severely hindered by scattering, but polarization imaging offers the unique dual advantages of descattering and shape-from-polarization (SfP) 3D reconstruction. To exploit these advantages, this paper proposes UD-SfPNet, an underwater descattering shape-from-polarization network that leverages polarization cues for improved 3D surface normal prediction. The framework jointly models polarization-based image descattering and SfP normal estimation in a unified pipeline, avoiding error accumulation from sequential processing and enabling global optimization across both tasks. UD-SfPNet further incorporates a novel color embedding module to enhance geometric consistency by exploiting the relationship between color encodings and surface orientation. A detail enhancement convolution module is also included to better preserve high-frequency geometric details that are lost under scattering. Experiments on the MuS-Polar3D dataset show that the proposed method significantly improves reconstruction accuracy, achieving a mean surface normal angular error of 15.12$^\circ$ (the lowest among compared methods). These results confirm the efficacy of combining descattering with polarization-based shape inference, and highlight the practical significance and potential applications of UD-SfPNet for optical 3D imaging in challenging underwater environments. The code is available at \url{https://github.com/WangPuyun/UD-SfPNet}.
\end{abstract}

\begin{graphicalabstract}
\begin{figure}[h]
	\centering
	\includegraphics[width=\columnwidth]{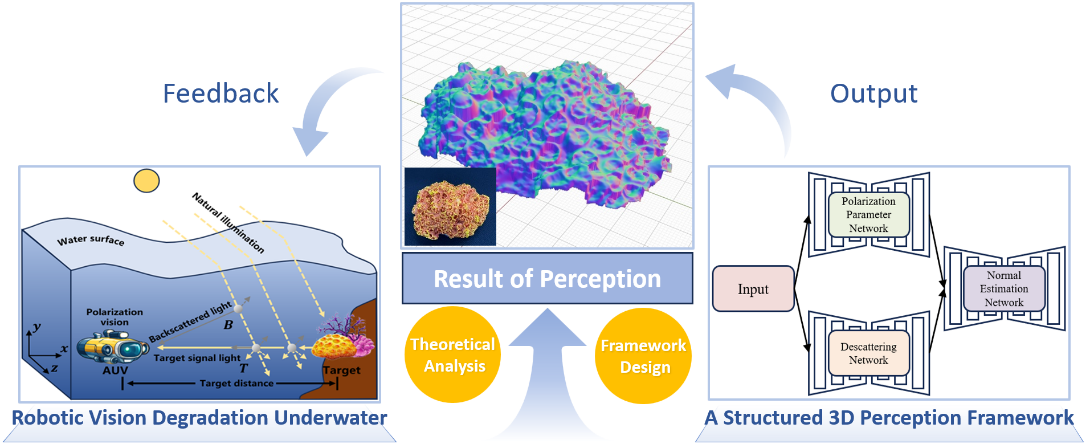}
	\caption{Schematic of the proposed underwater-robot polarization 3D visual perception pipeline.}
	\label{fig:Graphical_Abstract}
\end{figure}
\end{graphicalabstract}

\begin{highlights}
\item UD-SfPNet jointly trains underwater polarization dehazing and SfP normals
\item End-to-end training avoids error accumulation in cascade pipelines
\item Color-embedding transfer enforces geometry consistency via RGB-encoded normals
\item Detail-enhanced convs improve high-frequency geometry in both stages
\item 15.12° mean angular error on MuS-Polar3D, outperforming strong baselines
\end{highlights}

%% Keywords

\end{frontmatter}




\section{Introduction}
\label{sec1}
%% Labels are used to cross-reference an item using \ref command.

Underwater robots and manned submersibles rely on high-performance underwater imaging and 3D perception for ocean exploration. Optical 3D imaging in air has reached a high level of maturity. However, in scattering underwater environments, its performance degrades significantly due to Mie scattering in water~\cite{organelli2018open}. This degradation manifests as blurred texture details, a substantially reduced effective range, and severe noise. Motivated by computational imaging~\cite{xiang2024computational}, this paper rethinks underwater 3D imaging from the perspective of the imaging pipeline. The complete underwater 3D imaging pipeline is decomposed into two sequential stages: an initial descattering process followed by 3D reconstruction. Among various usable optical priors, light polarization shows unique advantages in both stages. Based on this observation, a full-chain optimization strategy is introduced: information is encoded across the decoupled descattering and reconstruction stages to push beyond conventional opto-electronic imaging limits, yielding results that surpass and improve upon the raw observations.

Most prior studies directly process underwater data with deep learning models~\cite{wu2025deep,li2025sfp}. Wang et al.~\cite{wang2025mus} explicitly formulate underwater 3D imaging as a two-stage pipeline consisting of descattering and subsequent 3D reconstruction. However, the two models are trained independently, and downstream vision tasks are performed on the descattered results, which may lead to error accumulation in cascaded processing pipelines. To address this issue, this paper proposes an underwater polarization-based 3D reconstruction method that integrates polarization physics and jointly models the descattering process and the 3D reconstruction process within a unified framework. The method enables global learning and joint optimization of upstream and downstream vision tasks, providing a new perspective for robotic underwater 3D visual perception.

The main contributions are as follows:
\begin{enumerate}
	\item This paper proposes UD-SfPNet, a unified structured learning framework for underwater polarization 3D imaging.
	The framework jointly learns polarization-based descattering and SfP normal estimation in an end-to-end manner,
	enabling global gradient optimization across modules.
	
	\item A color embedding module is introduced for SfP normal estimation.
	By exploiting the correspondence between RGB-encoded normal maps and geometric orientation,
	cross-channel consistency is enforced to enhance geometric stability.
	
	\item Detail-enhanced convolutions are incorporated in both the descattering and normal-estimation stages.
	These convolutions explicitly capture directional/differential variations,
	improving the recovery of high-frequency geometric details.
	
	\item Extensive experiments on MuS-Polar3D~\cite{wang2025mus} demonstrate the effectiveness of UD-SfPNet,
	achieving a mean angular error of 15.12$^{\circ}$ and outperforming strong baselines.
\end{enumerate}

\section{Related work}
\subsection{Polarization-based descattering}
The core idea of polarization-based descattering is to exploit the difference in polarization states between scattered light and object-reflected light, and separate these components via polarization computation to recover target information buried by scattering. In underwater scenes, photon scattering and absorption are stronger than in the atmosphere. As a result, polarization-based separation combined with attenuation compensation has become a predominant research direction. Early work by the Schechner group introduced polarization analysis for underwater visibility and structure recovery, and discussed the relation between backscatter polarization properties and the transmission function~\cite{schechner2006recovery}. The group later proposed active polarization descattering, which uses active illumination to enhance underwater polarization signals, reducing reliance on natural light and scanning hardware, and becoming one of the most representative physical models in this field~\cite{treibitz2008active}. With the recent progress of deep learning, this area has seen a new wave of advances. Learning-based descattering methods are typically grouped into two categories: physics-model-guided approaches and end-to-end enhancement approaches. Yang et al.~\cite{yang2024high} used multiple dilated convolutions to extract local polarization features around target pixels in polarization images, and combined a classical U-Net~\cite{ronneberger2015u} to recover high-quality polarization information for downstream tasks. Liu et al.~\cite{liu2024learning} focused more on end-to-end reconstruction and real-time performance under dynamic or strong-scattering conditions; by analyzing optical properties together with image-based evidence, the model was carefully tuned to achieve a descattering method with both real-time speed and robustness.

\subsection{Polarization-based 3D imaging}
Polarization-based 3D imaging, especially shape-from-polarization (SfP), is traditionally built on the Fresnel reflection model and Stokes parameters. Surface normals are inferred from the angle of polarization (AoP) and degree of polarization (DoP) by estimating the zenith and azimuth angles, and additional priors are used to further refine 3D reconstruction, as in~\cite{mahmoud2012direct,miyazaki2004transparent,smith2016linear}. However, such methods impose strict assumptions on material properties, leading to limited generalization.

In 2020, Ba et al.~\cite{ba2020deep} first introduced deep learning into SfP. A polarization dataset with complex materials and varying illumination was built for training, demonstrating that deep networks can adapt to diverse conditions for surface normal estimation and offering substantially higher potential than classical SfP. Later, Lei et al.~\cite{lei2022shape} extended SfP to outdoor complex scenes, Shao et al.~\cite{shao2023transparent} addressed 3D modeling difficulties caused by transmitted components in transparent objects, and Lyu et al.~\cite{lyu2024sfpuel} investigated the impact of unknown ambient illumination on SfP.

SfP tailored for underwater scattering has only been explored in the last two years. Wu et al.~\cite{wu2025deep} first applied SfP to underwater 3D target reconstruction. Based on U$^{2}$-Net~\cite{qin2020u2}, multi-scale features are extracted from raw polarization images and polarization-parameter maps, enabling accurate and robust 3D reconstruction in challenging underwater environments. Li et al.~\cite{li2025sfp} developed a higher-performance polarization 3D shape recovery network, SfP-underwater, using Transformers. However, it relies on a single encoder--decoder network to process polarization inputs and does not consider how an upstream task such as descattering affects the downstream task of 3D reconstruction. Wang et al.~\cite{wang2025mus} proposed explicit modeling of the descattering process for underwater SfP, but descattering and 3D reconstruction are handled by two separate deep models, with the output of the descattering model used as part of the reconstruction input. This cascaded design is also prone to error accumulation.

\section{Methodology}
\subsection{Principles of polarization imaging}

\begin{figure}[h]
	\centering
	\includegraphics[width=0.5\columnwidth]{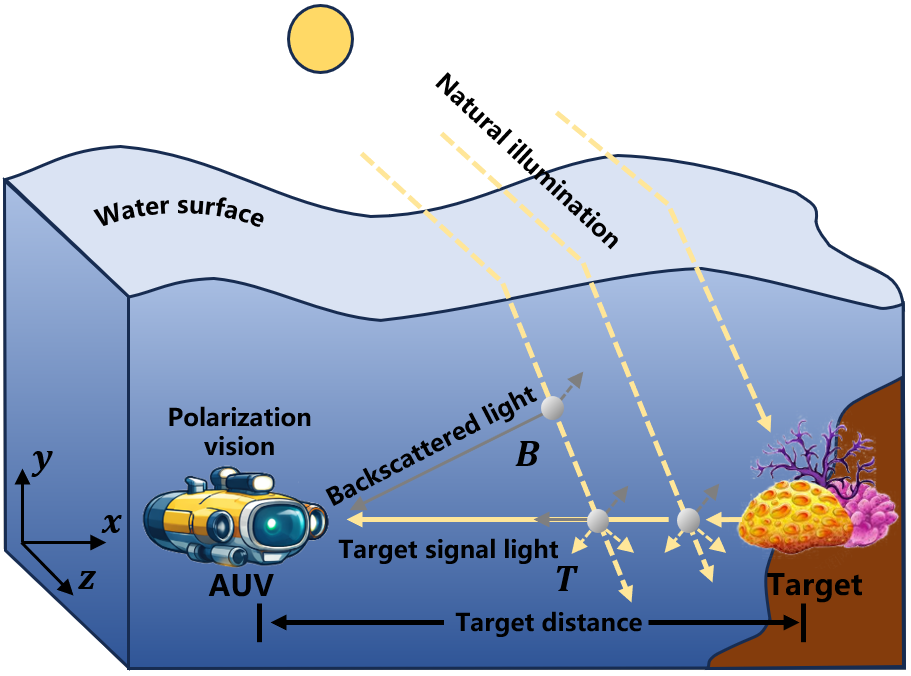}
	\caption{Underwater polarization imaging model.}
	\label{fig:Imaging model}
\end{figure}

In polarization computational imaging, the Stokes vector is widely used to represent and encode polarization characteristics~\cite{li2026monocular}. A division-of-focal-plane (DoFP) polarization camera~\cite{yamazaki2016four} can simultaneously capture four intensity images at different polarization angles, denoted as $I_{0^\circ}$, $I_{45^\circ}$, $I_{90^\circ}$, and $I_{135^\circ}$. Based on these four polarization images, the Stokes parameters can be computed as
\begin{equation}
	\mathbf{S}=
	\begin{bmatrix}
		S_0\\
		S_1\\
		S_2
	\end{bmatrix}
	=
	\begin{bmatrix}
		I_{0^\circ}+I_{90^\circ}\\
		I_{0^\circ}-I_{90^\circ}\\
		I_{45^\circ}-I_{135^\circ}
	\end{bmatrix}.
\end{equation}
Here, $S_0$ is the total intensity, $S_1$ is the difference between horizontal and vertical polarization components, and $S_2$ is the difference between the $45^\circ$ and $135^\circ$ components. According to the underwater polarization imaging model~\cite{yang2024high}, the captured signal can be further expressed as
\begin{equation}
	S_0(x,y)=T(x,y)+B(x,y),
\end{equation}
where $T(x,y)$ denotes the target signal and $B(x,y)$ denotes the backscattered light, with $(x,y)$ being pixel coordinates. Conventional opto-electronic imaging mainly relies on spectral and intensity cues. When a target is embedded in a background with similar radiance, its signal can be easily submerged. Polarization imaging exploits the polarization-state difference between backscattered light and target-reflected light, and separates the two components via polarization measurement to recover target information obscured by scattering. With strong nonlinear representation capability, deep learning can learn the intrinsic relation between polarization cues and scattering interference, enabling efficient optimization for polarization-based descattering.

\begin{figure}[h]
	\centering
	\includegraphics[width=0.5\columnwidth]{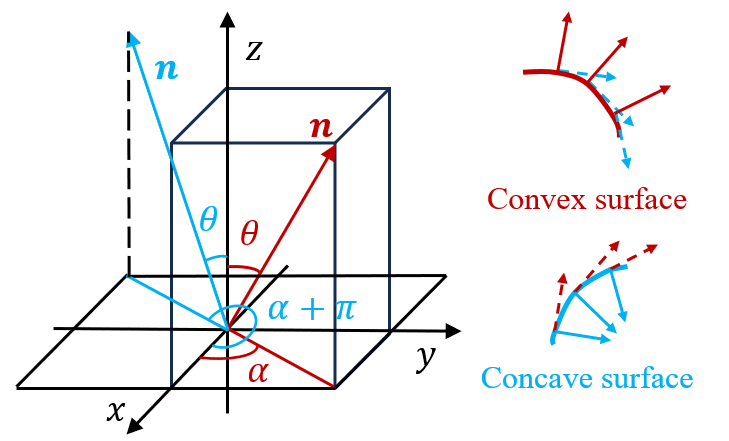}
	\caption{Schematic of the minimal element in the surface microfacet model.}
	\label{fig:Surface unit}
\end{figure}

Polarization-based 3D reconstruction essentially solves the zenith angle $\theta$ and azimuth angle $\alpha$ of the unit normal vector $\mathbf{n}$. As shown in Fig.~\ref{fig:Surface unit}, $\mathbf{n}$ can be expressed by $\theta$ and $\alpha$ as
\begin{equation}
	\mathbf{n}=
	\begin{bmatrix}
		\sin\theta\cos\alpha, \sin\theta\sin\alpha, \cos\theta
	\end{bmatrix}^\mathsf{T}.
	\label{eq: normal}
\end{equation}

Both the zenith angle $\theta$ and the azimuth angle $\alpha$ admit multiple solutions. For $\theta$, the ambiguity mainly comes from the nonlinear relation between the degree of polarization $\rho$ and $\theta$ under specular reflection. For $\alpha$, the ambiguity is caused by the periodicity of the sine function. This periodicity yields two ambiguous solutions whose phases differ by $\pi$, which can further make local normal reconstruction appear as either a ``convex'' or ``concave'' surface (see Fig.~\ref{fig:Surface unit}).

\begin{figure}[h]
	\centering
	\includegraphics[width=\columnwidth]{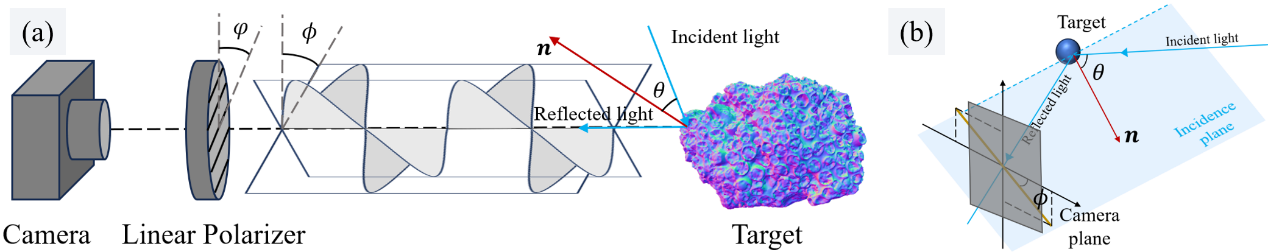}
	\caption{Physical model of polarization-based 3D reconstruction: (a) global view; (b) camera view.}
	\label{fig:3D imaging model}
\end{figure}

Conventional polarization-based 3D reconstruction is typically built on the physical model in Fig.~\ref{fig:3D imaging model}. The basic idea is to compute the degree of polarization $\rho$ and the angle of polarization $\phi$, and then invert the zenith angle $\theta$ and azimuth angle $\alpha$ of the unit surface normal. The degree of polarization $\rho$ and the angle of polarization $\phi$ can be obtained from the Stokes parameters as
\begin{equation}
	\rho=\frac{\sqrt{S_1^2+S_2^2}}{S_0},
\end{equation}
\begin{equation}
	\phi=\frac{1}{2}\arctan\left(\frac{S_2}{S_1}\right).
\end{equation}

\subsubsection{Solving the zenith angle $\theta$}
\begin{figure}[h]
	\centering
	\includegraphics[width=0.8\columnwidth]{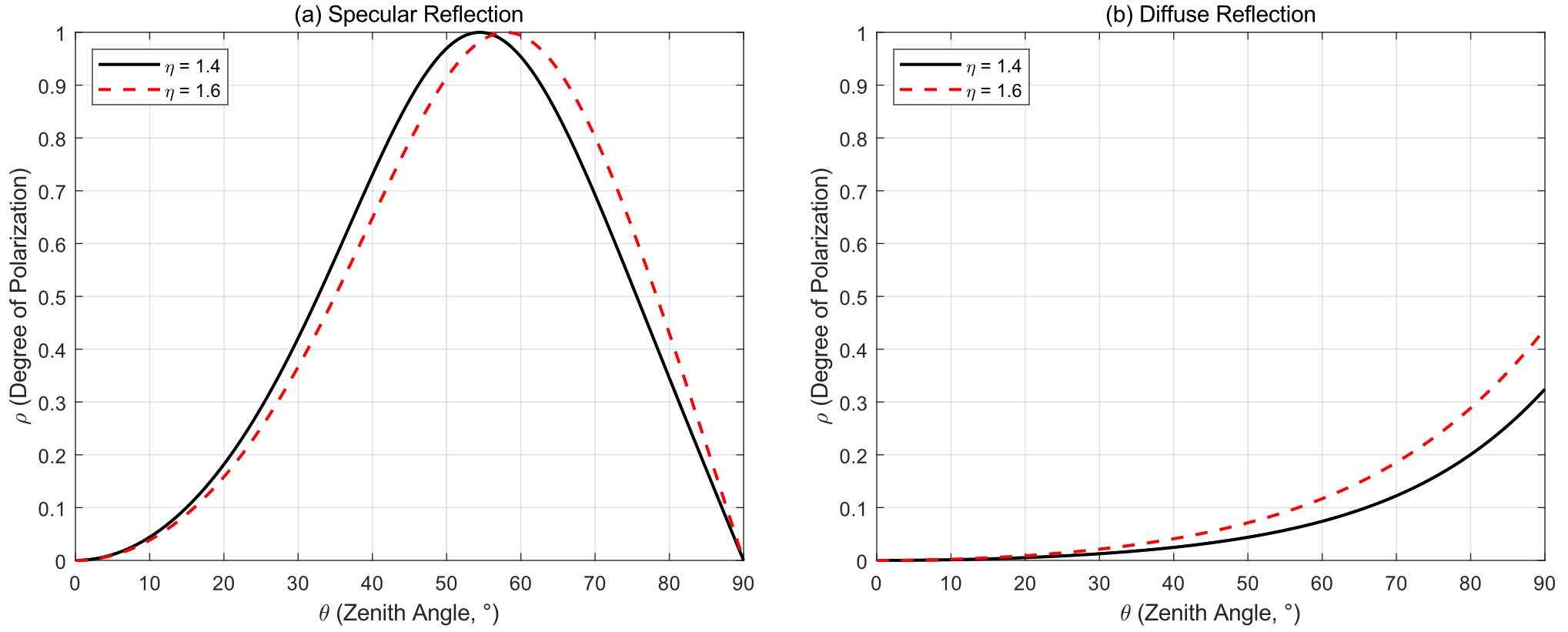}
	\caption{Relation between the degree of polarization $\rho$ and the zenith angle $\theta$ for a given refractive index $\eta$: (a) specular reflection; (b) diffuse reflection.}
	\label{fig:Zenith2DoP}
\end{figure}

Given the refractive index $\eta$ of the target surface, under specular reflection the relation between the degree of polarization $\rho$ and the zenith angle $\theta$ is
\begin{equation}
	\rho=\frac{2\sin^2\theta\cos\theta\sqrt{\eta^2-\sin^2\theta}}{\eta^2-\sin^2\theta-\eta^2\sin^2\theta+2\sin^4\theta}.
\end{equation}
The corresponding curve is shown in Fig.~\ref{fig:Zenith2DoP}(a). When and only when $\rho=1$, the zenith angle equals the Brewster angle, and $\theta$ is unique. Otherwise, when $\rho\neq 1$, $\theta$ has two distinct solutions.

For diffuse reflection, the relation between the degree of polarization $\rho$ and the zenith angle $\theta$ is given by
\begin{equation}
	\rho=\frac{\left(\eta-\frac{1}{\eta}\right)^{2}\sin^{2}\theta}{2+2\eta^{2}-\left(\eta+\frac{1}{\eta}\right)^{2}\sin^{2}\theta+4\cos\theta\sqrt{\eta^{2}-\sin^{2}\theta}},
\end{equation}
and the corresponding trend is shown in Fig.~\ref{fig:Zenith2DoP}(b). With the above two relations, the zenith angle $\theta$ can be solved from the measured $\rho$.

\subsubsection{Solving the azimuth angle $\alpha$}
\begin{figure}[h]
	\centering
	\includegraphics[width=0.5\columnwidth]{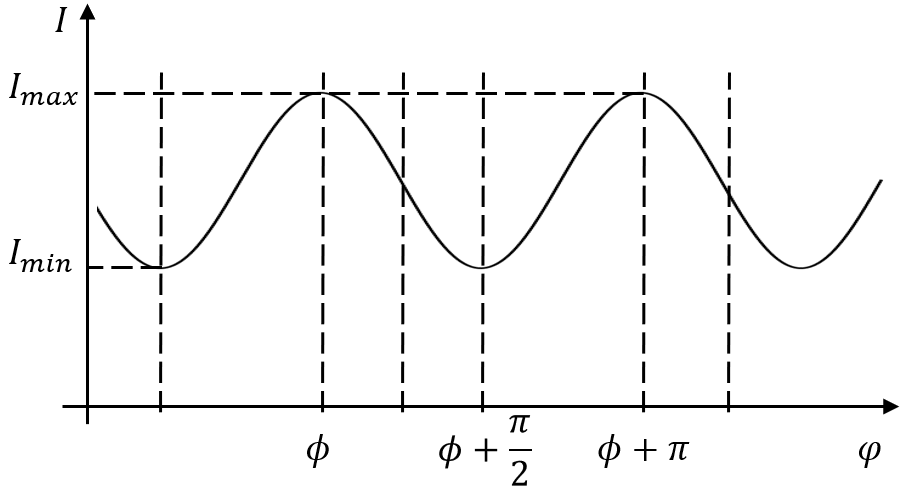}
	\caption{Relation between the intensity $I$ and the polarizer rotation angle $\varphi$.}
	\label{fig:Angle2light}
\end{figure}

The captured intensity $I$ varies with the polarizer rotation angle $\varphi$ as
\begin{equation}
	I(\varphi)=\frac{I_{\max}+I_{\min}}{2}+\frac{I_{\max}-I_{\min}}{2}\cos\left(2\varphi-2\phi\right).
\end{equation}
The corresponding trend is shown in Fig.~\ref{fig:Angle2light}. When $\varphi=\phi$, the above equation gives $I(\phi)=I_{\max}$. In this case, the angle between the plane of incidence and the coordinate axes of the camera image plane (i.e., the AoP $\phi$, as in Fig.~\ref{fig:3D imaging model}(b)) equals the polarizer rotation angle $\varphi$. For diffuse reflection, the azimuth angle satisfies $\alpha=\phi$ or $\alpha=\phi+\pi$. For specular reflection, $\alpha=\phi\pm\frac{\pi}{2}$. When $\varphi=\phi\pm\frac{\pi}{2}$, $I(\varphi)=I_{\min}$; this property is also used to suppress specular glare from objects such as glass.

With $\phi$, the azimuth angle $\alpha$ can be obtained. Together with the zenith angle $\theta$ from the previous subsection, substituting $\theta$ and $\alpha$ into equation~\ref{eq: normal} yields the unit normal of a surface element. Integrating the normal field over the surface then provides the full 3D shape of the target.

Conventional polarization-based 3D imaging often requires auxiliary sensors (e.g., depth cameras) to provide extra physical cues for resolving ambiguities in the zenith and azimuth angles. In contrast, deep learning can use supervised training with ground-truth annotations to let the model learn polarization physics implicitly. Although normal ambiguities cannot be completely eliminated, learning-based methods show strong generalization to such ambiguities, e.g., delivering accurate normal predictions on MuS-Polar3D samples with diverse materials and optical properties.

\subsection{Network architecture}
\begin{figure*}[!h]
	\centering
	\includegraphics[width=\textwidth]{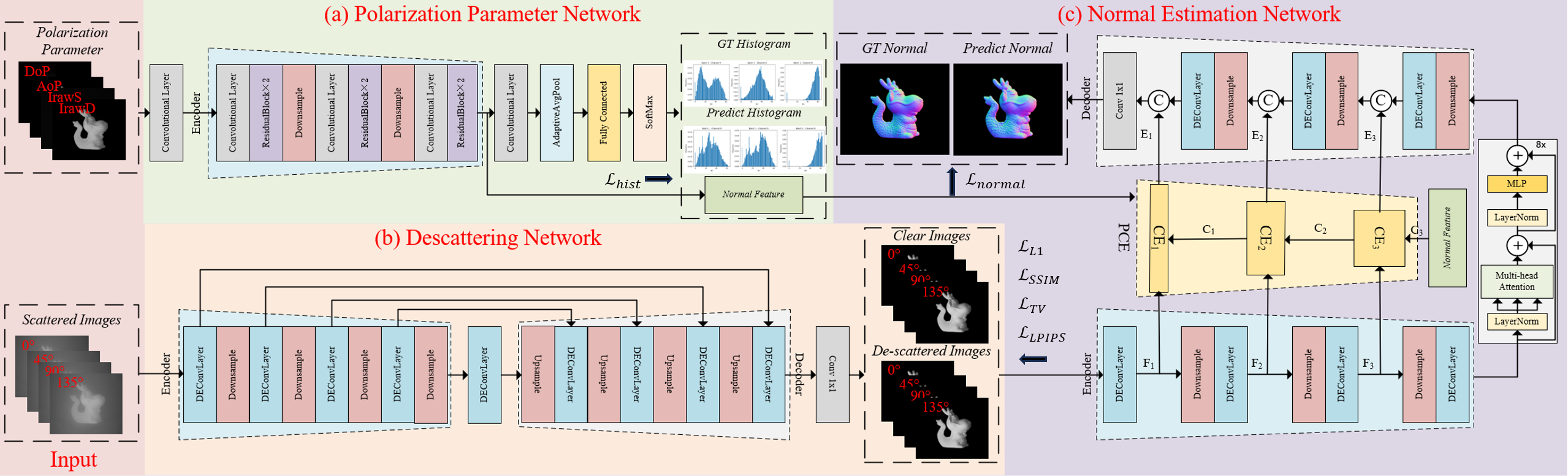}
	\caption{Overview of the structured learning framework UD-SfPNet for underwater polarization-based 3D reconstruction.}
	\label{fig:Network}
\end{figure*}

This section introduces the proposed structured learning framework UD-SfPNet (Fig.~\ref{fig:Network}) and its architectural details. The goal is to fuse polarization cues for 3D reconstruction in underwater scattering environments. The framework is composed of three parts: (a) a polarization-parameter network, (b) a descattering network, and (c) a normal-estimation network.

\subsubsection{Polarization Parameter Network (PPN)}
Given polarization inputs, this part trains a network to learn the mapping from polarization features to 3D surface normals. The polarization input combination described in~\cite{wu2023three} is adopted, and the network is built as an ``image encoder + classification head''. The output of the classification head is supervised by a normal-distribution histogram computed from the ground-truth normal map. Only the global proportion of normals is learned (64 bins), while the spatial distribution of each normal is not considered at this stage. During training, the high-dimensional normal feature predicted by the encoder (normal feature, NF) is retained for subsequent 3D reconstruction. The process is formulated as
\begin{equation}
	\mathrm{NF}_{\mathrm{pre}},\ \mathrm{Hist}_{\mathrm{pre}}
	=
	\mathrm{PPN}\!\left(\rho,\ \phi,\ I_{\mathrm{raw}}^{S},\ I_{\mathrm{raw}}^{D}\right),
\end{equation}
where $\mathrm{NF}_{\mathrm{pre}}$ is the predicted normal feature, $\mathrm{Hist}_{\mathrm{pre}}$ is the predicted histogram over 64 normal bins, and $\mathrm{PPN}(\cdot)$ denotes the polarization-parameter network. Here, $\rho$ is the degree of polarization, $\phi$ is the angle of polarization, $I_{\mathrm{raw}}^{S}$ is the specular component, and $I_{\mathrm{raw}}^{D}$ is the diffuse component, all computed from the raw polarization images. An $\ell_{1}$ constraint is used:
\begin{equation}
	\mathcal{L}_{\mathrm{hist}}
	=
	\left\|
	\mathrm{Hist}_{\mathrm{pre}}-\mathrm{Hist}_{\mathrm{GT}}
	\right\|_{1},
\end{equation}
where $\mathcal{L}_{\mathrm{hist}}$ is the histogram loss and $\mathrm{Hist}_{\mathrm{GT}}$ is the ground-truth histogram computed from the normal map. This loss constrains the classification head and indirectly optimizes the image encoder, enabling learning from polarization cues toward normal prediction.

\subsubsection{Descattering Network (DN)}
The descattering network focuses on a low-level vision task. Its goal is to enhance the contrast and sharpness of polarization images under scattering, and recover target information submerged by scattering. The network follows a classical U-Net~\cite{ronneberger2015u} design, which is an encoder--decoder with four levels of skip connections. The overall process is
\begin{equation}
	I_{\mathrm{desc}}=\mathrm{DN}(I_{\mathrm{sc}}),
\end{equation}
where $I_{\mathrm{sc}}$ is the input scattered image, $\mathrm{DN}(\cdot)$ denotes the descattering operation, and $I_{\mathrm{desc}}$ is the descattered output. Four losses are used for supervision:
\begin{equation}
	\mathcal{L}_{\mathrm{L1}}=\left\|I_{\mathrm{desc}}-I_{\mathrm{clean}}\right\|_{1},
\end{equation}
\begin{equation}
	\mathcal{L}_{\mathrm{SSIM}}=1-\mathrm{SSIM}\!\left(I_{\mathrm{desc}},I_{\mathrm{clean}}\right),
\end{equation}
\begin{equation}
	\mathrm{SSIM}(x,y)=\frac{\left(2\mu_x\mu_y+C_1\right)\left(2\sigma_{xy}+C_2\right)}{\left(\mu_x^{2}+\mu_y^{2}+C_1\right)\left(\sigma_x^{2}+\sigma_y^{2}+C_2\right)},
\end{equation}
\begin{equation}
	\mathcal{L}_{\mathrm{TV}}=\sum_{i,j}\left(\left|I_{i+1,j}-I_{i,j}\right|+\left|I_{i,j+1}-I_{i,j}\right|\right),
\end{equation}
\begin{equation}
	\mathcal{L}_{\mathrm{LPIPS}}=\mathrm{LPIPS}\!\left(I_{\mathrm{mean}}^{\mathrm{desc}},I_{\mathrm{mean}}^{\mathrm{clean}}\right).
\end{equation}
Here, $I_{\mathrm{clean}}$ denotes the polarization image captured in clear water and is used as the reference. $\|\cdot\|_{1}$ is the $\ell_{1}$ norm to enforce pixel-wise reconstruction accuracy. SSIM is the structural similarity index, which constrains consistency between the descattered result and the reference in terms of luminance, contrast, and structure; $\mu$ and $\sigma$ denote mean and variance in the SSIM computation. $\mathcal{L}_{\mathrm{TV}}$ is the total variation loss, which promotes spatial smoothness and suppresses high-frequency noise and local artifacts introduced by scattering. $\mathcal{L}_{\mathrm{LPIPS}}$ is a perceptual loss; prior work~\cite{zhang2018unreasonable} shows that it helps improve the perceptual naturalness of descattering results. Since LPIPS is trained mainly on 3-channel natural images, the four polarization channels are first converted into a mean-intensity image and then replicated to three channels as the LPIPS input. $I_{\mathrm{mean}}^{\mathrm{desc}}$ and $I_{\mathrm{mean}}^{\mathrm{clean}}$ denote the mean-intensity maps of the descattered output and the reference, respectively.

\subsubsection{Normal Estimation Network (NEN)}
The normal estimation network performs the high-level 3D reconstruction task. It integrates the outputs from the descattering network and the polarization parameter network, and jointly recovers high-accuracy surface normals. The architecture contains one encoder and two decoders. A multi-head attention module~\cite{vaswani2017attention} is used at the bottleneck to improve high-dimensional feature aggregation. Notably, one decoder adopts the Pyramid Color Embedding (PCE) module proposed in~\cite{zhang2022deep}. The color-encoding property of PCE matches the normal-encoding rule in SfP (introduced in Sec.~\ref{subsection: Color embedding module}). In essence, it is a feature integration and lifting of low-level bio-visual cues. Therefore, this paper extends it from 2D low-light enhancement to 3D robotic visual perception, leading to strong reconstruction accuracy. The normal estimation is formulated as
\begin{equation}
	N_{\mathrm{pre}}=\mathrm{NEN}\!\left(\mathrm{NF},I_{\mathrm{desc}}\right),
\end{equation}
where $N_{\mathrm{pre}}$ is the predicted normal map and $\mathrm{NEN}(\cdot)$ denotes the overall normal-estimation process. The loss is defined as
\begin{equation}
	\mathcal{L}_{\mathrm{normal}}
	=
	\frac{
		\sum_{i=1}^{W}\sum_{j=1}^{H}
		\left(1-\left\langle N_{i,j}^{\mathrm{pre}},N_{i,j}^{\mathrm{gt}}\right\rangle\right)-m
	}{
		W\times H-m
	},
\end{equation}
where $\langle\cdot,\cdot\rangle$ denotes cosine similarity, $W$ and $H$ are the image width and height, and $m$ is the number of background pixels. This loss measures angular deviation via cosine similarity and uses a mask to normalize only over the foreground region, improving stability and accuracy of normal prediction.

\begin{figure}[!ht]
	\centering
	\includegraphics[width=0.8\columnwidth]{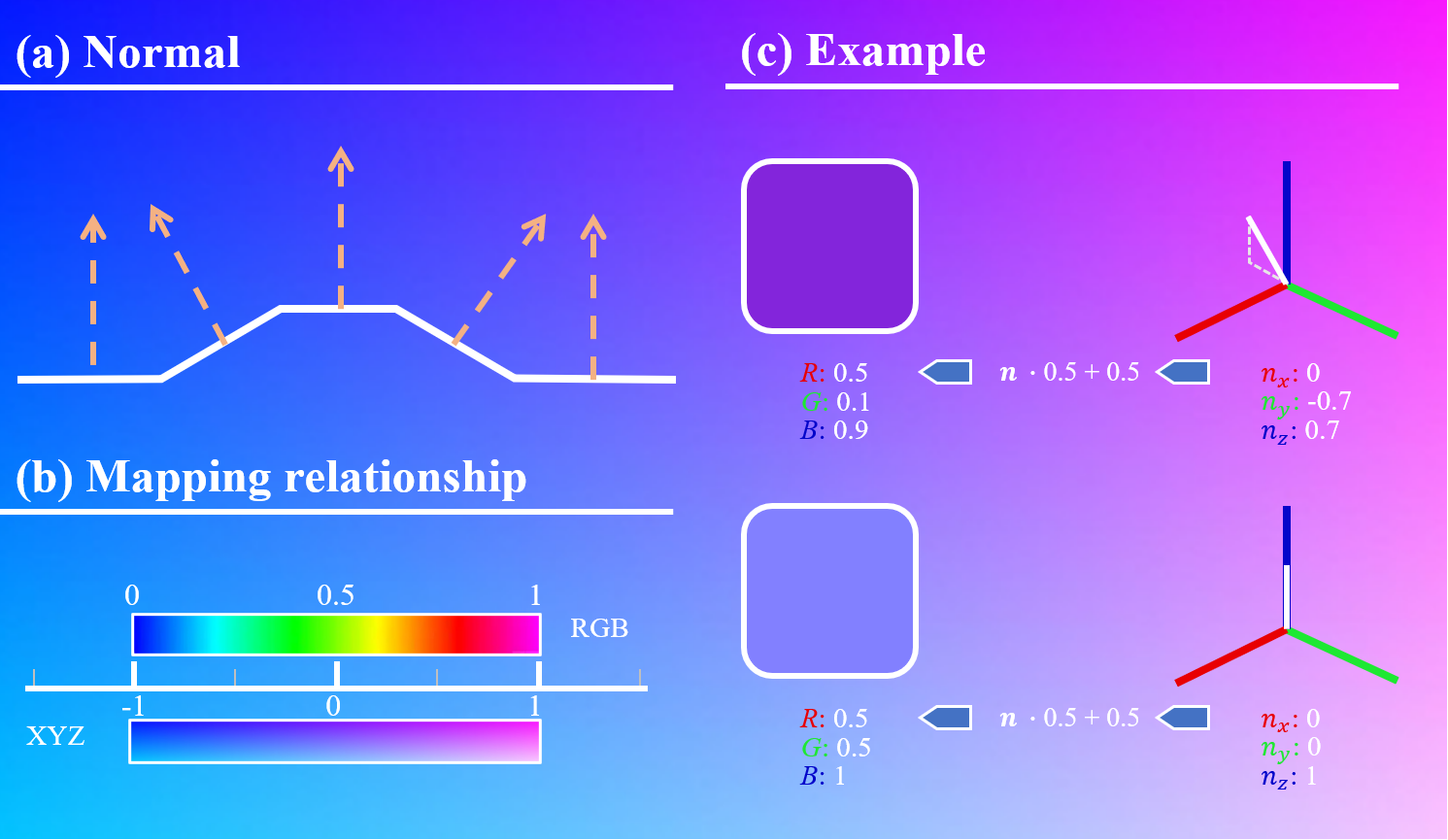}
	\caption{Color encoding mechanism of baked normal maps.}
	\label{fig:Color_Encoding_Mechanism}
\end{figure}

\subsection{Color embedding module}
\label{subsection: Color embedding module}
This subsection describes the color embedding module from DCC~\cite{zhang2022deep} and its potential in SfP. In SfP, surface normals are commonly represented and supervised using a normal map. As shown in Fig.~\ref{fig:Color_Encoding_Mechanism}, this representation is essentially a color-based vector encoding: for each pixel, the unit normal $\mathbf{n}=\left(n_x,n_y,n_z\right)^\mathsf{T}$ is mapped to the RGB channels, allowing 3D geometry to be compactly stored on a 2D image plane. More specifically, a standard normal map applies linear normalization from $[-1,1]$ to $[0,1]$, yielding the color code
\begin{equation}
	\mathbf{c}=\left(R,G,B\right)^{\mathrm{T}}=\frac{\mathbf{n}+1}{2},\ \mathrm{where}\ R\leftrightarrow n_x,\ G\leftrightarrow n_y,\ B\leftrightarrow n_z.
\end{equation}
This encoding is widely used in graphics normal baking and in normal-supervised learning, making 3D normal prediction equivalent to recovering a ``color image'' with strict geometric constraints.

Based on this observation, SfP normal prediction can be reframed as a representation learning problem driven by color consistency. If the network learns a degradation-robust and consistent color embedding in feature space, this consistency can be naturally transferred to normal-direction consistency through the RGB encoding of the normal map, improving the stability and usability of geometric recovery.

The Color Embedding module proposed in DCC~\cite{zhang2022deep} was designed to handle color shift under low illumination. Its key idea is to structurally model the color distribution in feature space so that the output preserves color consistency in both spatial and semantic senses. Although originally developed for low-light enhancement, the core challenge it addresses---maintaining stable and consistent color under strong degradation---matches the encoding property of normal maps in underwater scattering SfP.

Following the above analysis, the Color Embedding module is adapted for SfP normal prediction in underwater scattering environments. This is not a naive plug-in; it follows directly from the ``color--geometry'' isomorphism of normal maps, where color consistency implies geometric consistency.

\begin{figure}[!ht]
	\centering
	\includegraphics[width=0.6\columnwidth]{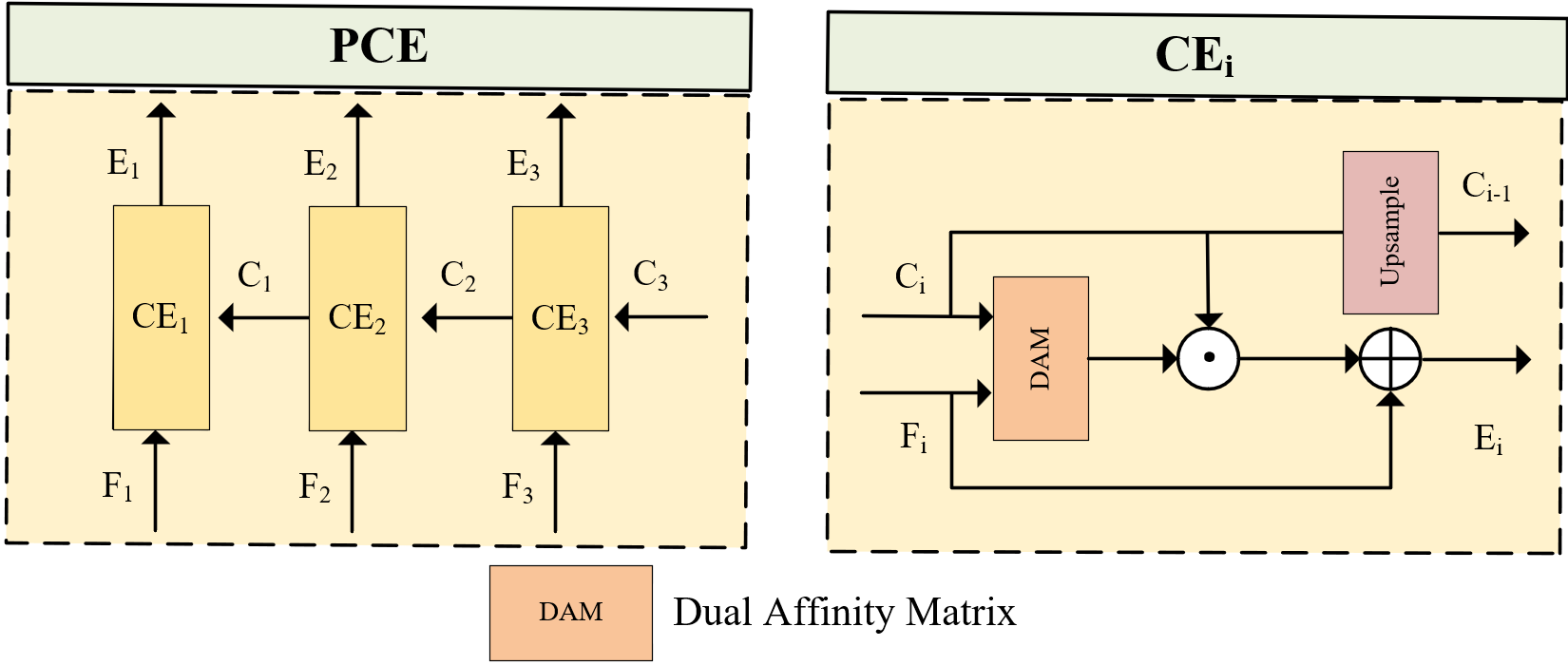}
	\caption{Color embedding module adapted from DCC~\cite{zhang2022deep}: the pyramid color embedding (PCE) composed of multi-level color embedding modules (left), and the detailed implementation of the color embedding (CE) module (right).}
	\label{fig:Color_Embedding_Module}
\end{figure}

\subsection{Detail-enhanced convolution module}

\begin{figure}[!ht]
	\centering
	\includegraphics[width=0.7\columnwidth]{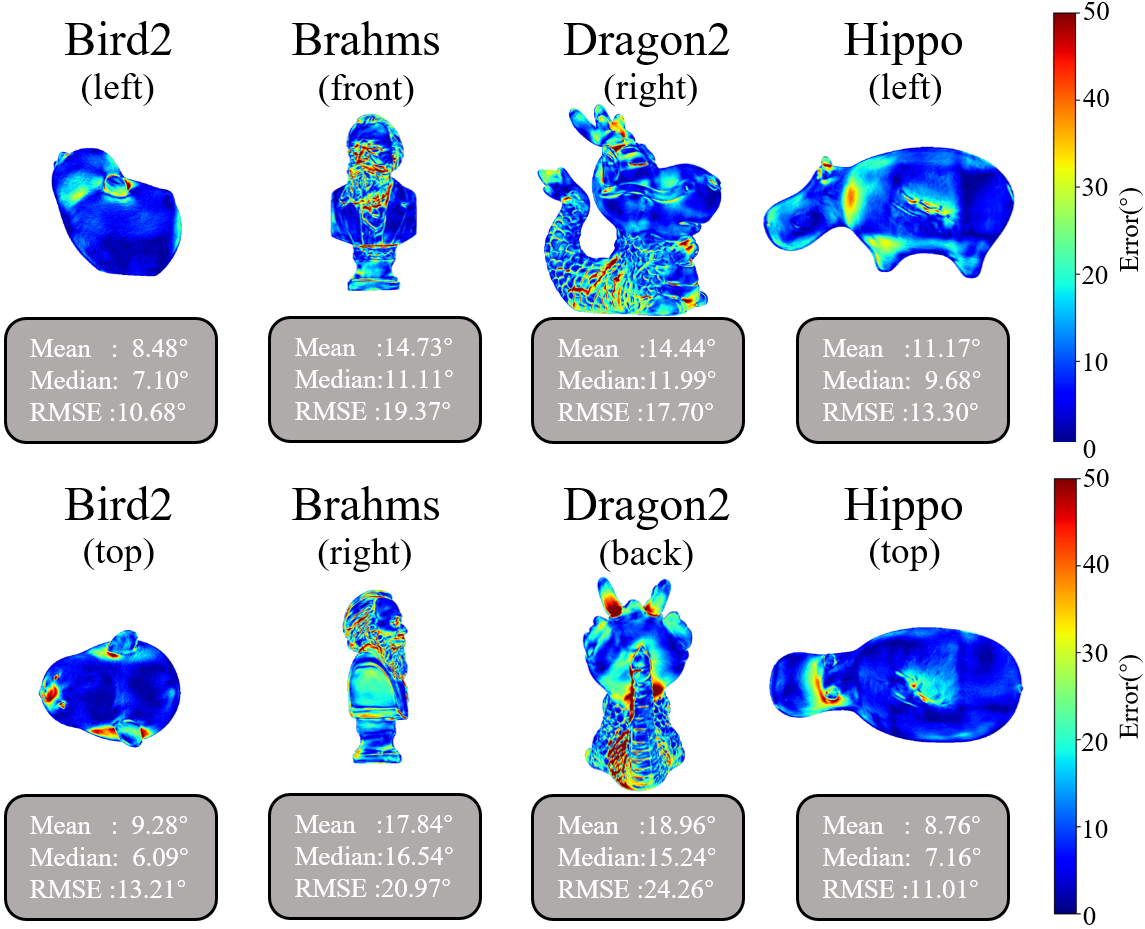}
	\caption{Comparison of error heatmaps across different samples.}
	\label{fig:heatmap}
\end{figure}

As shown in Fig.~\ref{fig:heatmap}, Wang et al.~\cite{wang2025mus} used error heatmap visualization to show that, once descattering effectively suppresses scattering degradation, the main bottleneck of 3D reconstruction shifts to the intrinsic geometry of the target. In particular, regions with richer high-frequency geometric textures tend to exhibit larger normal-estimation and reconstruction errors.

This indicates that the key challenge of normal estimation lies in accurate modeling of high-frequency geometric details. However, standard convolution operators mainly aggregate overall responses within a local neighborhood and have limited sensitivity to local differential cues and directional variations, which can restrict prediction accuracy in detail-rich regions.

Based on this observation, the Detail-Enhanced Convolution (DEConv) module from DEA-Net~\cite{chen2024dea} is introduced as the convolution backbone in both the low-level descattering network and the high-level normal-estimation network. Built upon standard convolution, this module incorporates multiple differential convolution operators to explicitly model local pixel differences and directional changes, thereby strengthening the representation of high-frequency structural information.

Introducing DEConv in both key tasks, i.e., ``descattering'' and ``3D reconstruction'', improves the overall network stability and reconstruction accuracy in complex geometric regions.

\begin{figure}[!ht]
	\centering
	\includegraphics[width=0.5\columnwidth]{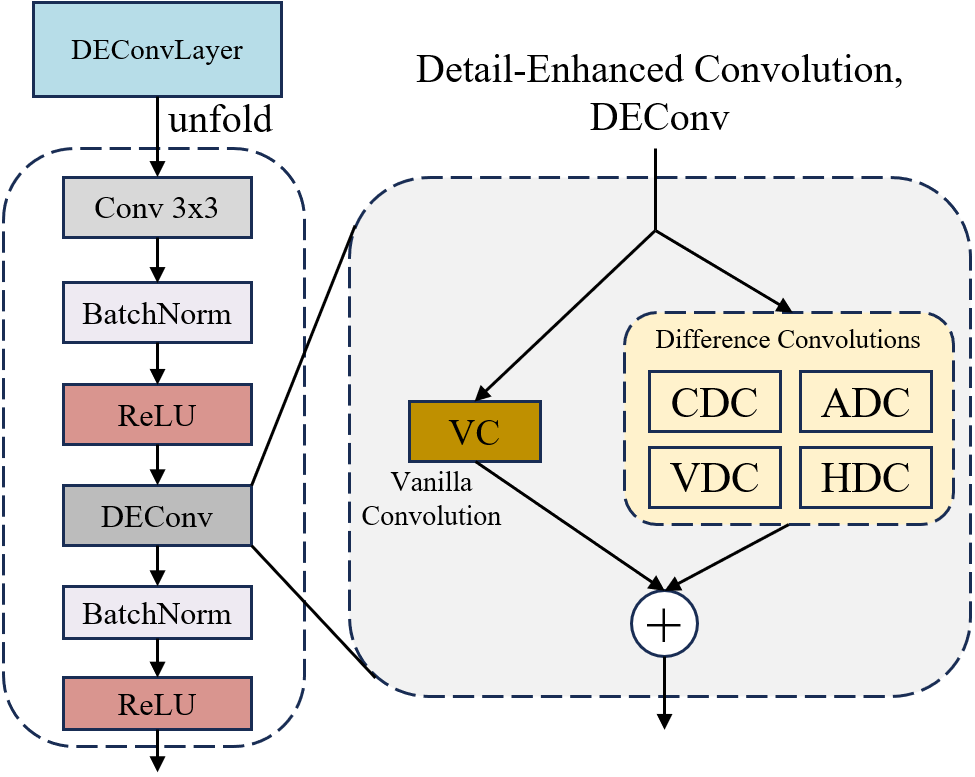}
	\caption{Schematic illustration of the DEConv module.}
	\label{fig:DEConv}
\end{figure}

\subsection{Further discussion}
The proposed structured learning framework for underwater polarization-based 3D reconstruction explicitly models two processes in underwater 3D imaging, namely ``descattering'' and ``3D reconstruction'', enabling global optimization during deep learning training. In addition, each sub-network can be replaced by alternative methods to further improve performance. For example, Transformer-based models have been applied to scattering removal~\cite{song2023vision} and surface normal estimation~\cite{wan2025attentivesfp}. Transformers typically require large-scale training data. Within this structured framework, other components can be frozen so that a target sub-network is pretrained on large public descattering/normal-estimation datasets and then fine-tuned on a specific dataset for best underwater 3D imaging performance. If real-time performance is required, complex modules can be removed and replaced with standard convolution blocks to obtain a lightweight implementation.

\section{Experiments and results}
\subsection{Training setup}
The 726 scattering samples from the public MuS-Polar3D dataset~\cite{wang2025mus} are split into training/validation/test sets with an 8:1:1 ratio. The proposed 3D imaging framework is trained for 1000 epochs on the PyTorch~\cite{paszke2019pytorch} platform using 4 NVIDIA A100 GPUs with a learning rate of 0.001. During training, $256\times256$ patches are randomly cropped with more than half of the pixels being valid, and patch-based training is used for data augmentation. Inference is also performed on $256\times256$ patches; a sliding-window strategy is used to stitch patches back to full-resolution images and reduce block artifacts. The total loss is
\begin{equation}
	\mathcal{L}_{\mathrm{total}}
	=
	\lambda_1\mathcal{L}_{\mathrm{hist}}
	+\lambda_2\mathcal{L}_{\mathrm{L1}}
	+\lambda_3\mathcal{L}_{\mathrm{SSIM}}
	+\lambda_4\mathcal{L}_{\mathrm{TV}}
	+\lambda_5\mathcal{L}_{\mathrm{LPIPS}}
	+\lambda_6\mathcal{L}_{\mathrm{normal}}.
\end{equation}
The hyperparameters are set empirically to $\lambda_1=1.0$, $\lambda_2=10.0$, $\lambda_3=1.0$, $\lambda_4=10.0$, $\lambda_5=2.0$, and $\lambda_6=30.0$.

\subsection{Performance evaluation}
\begin{figure*}[!ht]
	\centering
	\includegraphics[width=\columnwidth]{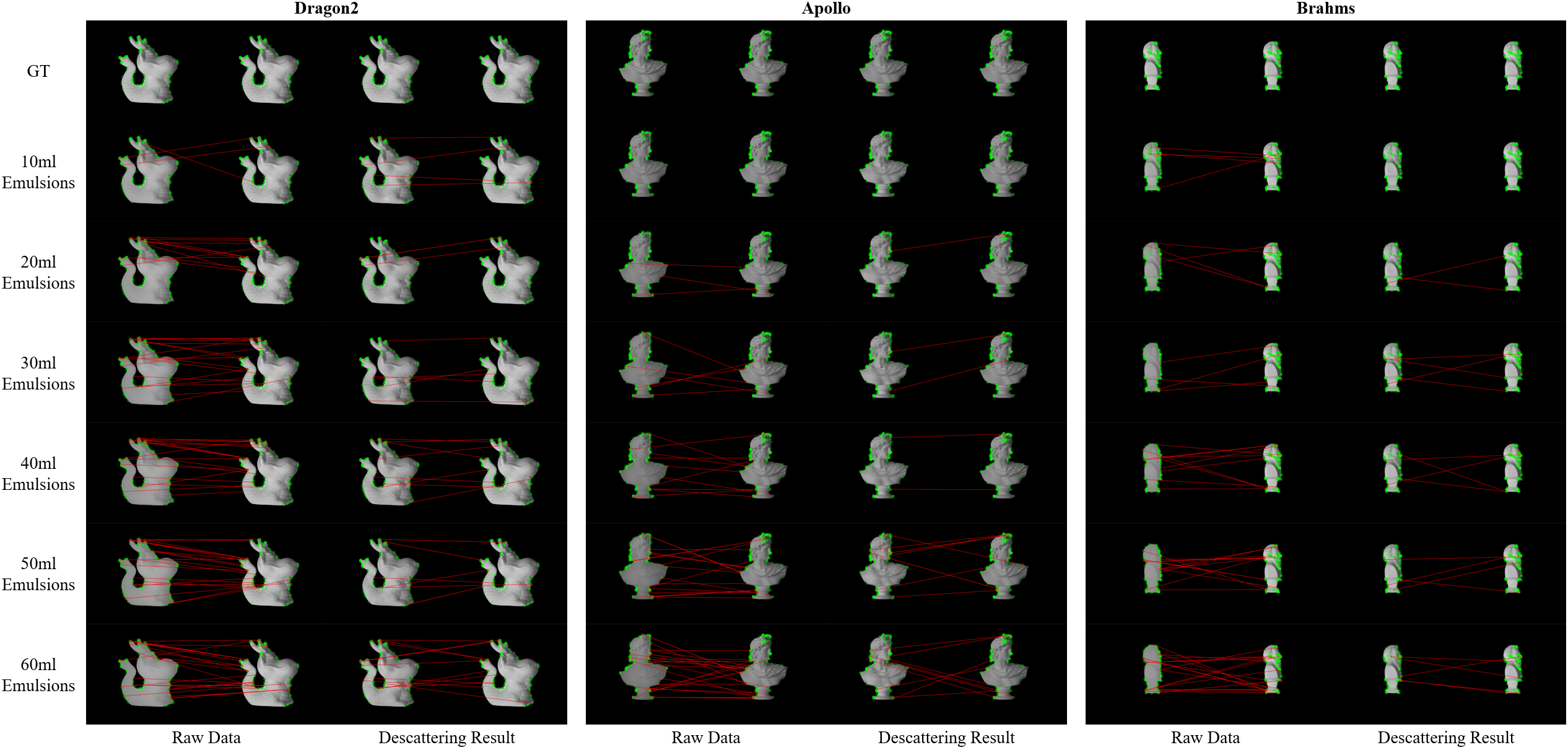}
	\caption{Descattering evaluation using the false match rate of ORB-based registration. Green circles indicate detected keypoints, and red lines indicate incorrect matches.}
	\label{fig:ORB}
\end{figure*}

The proposed SfP method explicitly models the descattering process in underwater environments. This subsection first evaluates descattering performance. As shown in Fig.~\ref{fig:ORB}, qualitative evaluation is conducted on representative samples using the false match rate of the ORB~\cite{rublee2011orb} registration algorithm. After processing by the descattering network, the number of red lines indicating incorrect correspondences decreases markedly, suggesting that the descattering network is effective.

Table~\ref{tab:descattering_metrics} further compares the raw images, the descattered results, and the ground-truth references. Peak signal-to-noise ratio (PSNR), structural similarity (SSIM), and learned perceptual image patch similarity (LPIPS) are reported. To better reflect visual enhancement in target regions, masks are applied to both the ground truth and the evaluated samples during metric computation. The results show that, after descattering, all metrics improve over the target region, indicating that descattering enhances visual quality and can provide richer effective information for downstream tasks.

\begin{table}[t]
	\centering
	\caption{Quantitative comparison before and after descattering. Masks are applied to both the ground truth and the evaluated samples to focus on the target region.}
	\label{tab:descattering_metrics}
	\begin{tabular}{lccc}
		\hline
		Metric & PSNR$\uparrow$ & SSIM$\uparrow$ & LPIPS$\downarrow$ \\
		\hline
		Raw data & 30.80 & 0.9569 & 0.3830 \\
		Descattering result & 36.87 & 0.9745 & 0.0356 \\
		\hline
	\end{tabular}
\end{table}

\begin{figure*}[!ht]
	\centering
	\includegraphics[width=\textwidth]{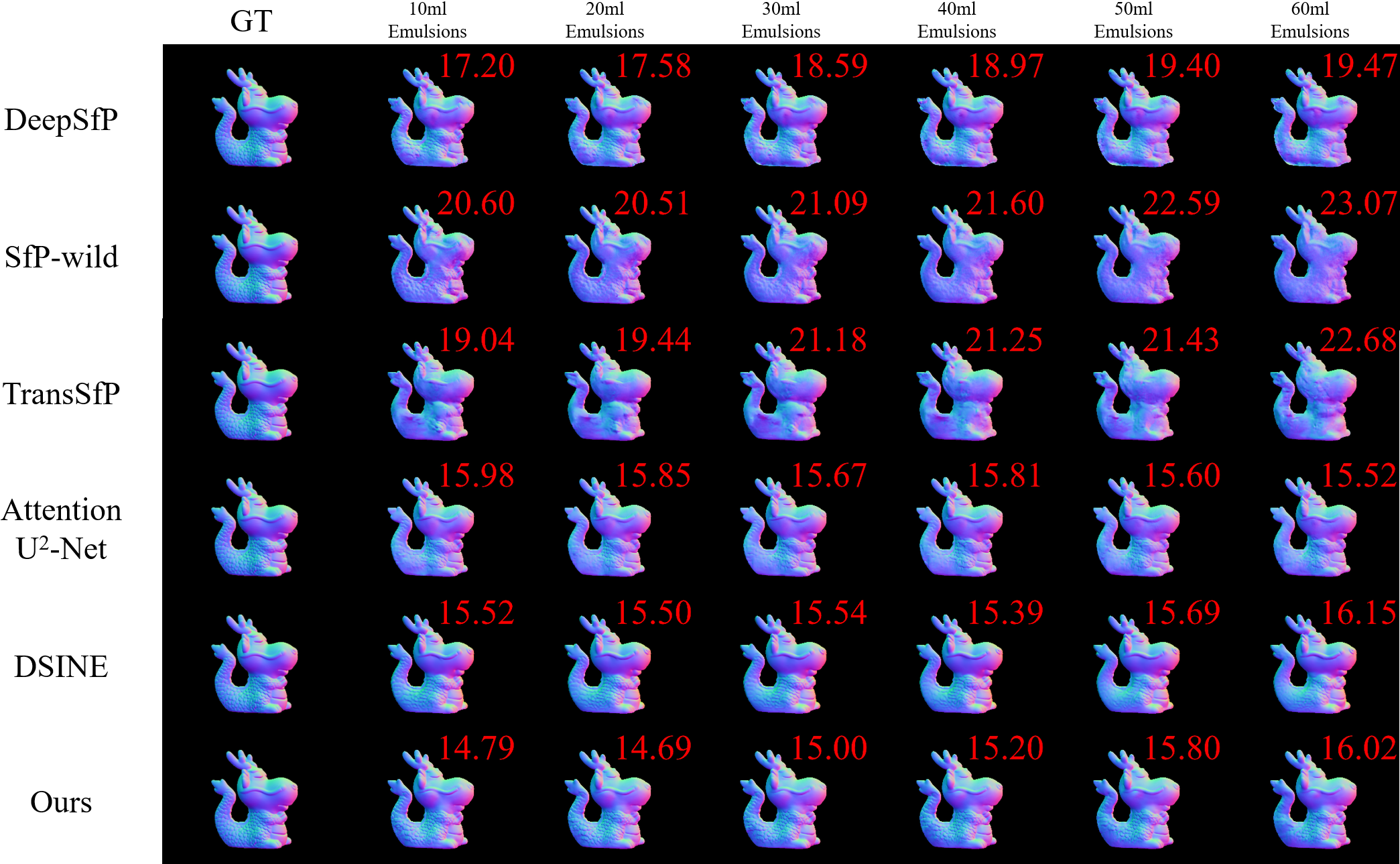}
	\caption{Comparison of different baseline methods on a representative sample (Dragon2). The red numbers in the upper-right corner of each subfigure denote the MAE (°), where lower values indicate better performance.}
	\label{fig:Baseline_Compare}
\end{figure*}

The proposed method is evaluated against five normal estimation baselines, including DeepSfP~\cite{ba2020deep}, SfP-wild~\cite{lei2022shape}, TransSfP~\cite{shao2023transparent}, AttentionU$^{2}$-Net~\cite{wu2025deep}, and DSINE~\cite{bae2024rethinking}. Table~\ref{tab:mae_baselines} reports the mean angular error (MAE) on the MuS-Polar3D test set. The proposed method achieves the lowest MAE, supporting the effectiveness of jointly modeling low-level vision (descattering) and high-level vision (3D reconstruction). Fig.~\ref{fig:Baseline_Compare} further visualizes predictions on representative samples. DeepSfP, SfP-wild, and TransSfP are more affected by underwater scattering, showing increasing MAE as scattering becomes stronger.  AttentionU$^{2}$-Net, DSINE, and the proposed method are more robust to underwater conditions, with the proposed method yielding the lowest MAE.

\begin{table}[t]
	\centering
	\caption{Performance comparison of different baselines on the MuS-Polar3D test set. The metric is mean angular error (MAE, $^\circ$); lower is better.}
	\label{tab:mae_baselines}
	\begin{tabular}{lcccccc}
		\hline
		Metric & DeepSfP & SfP-wild & TransSfP & \begin{tabular}{@{}c@{}}Attention\\U2-Net\end{tabular} & DSINE & Proposed \\
		\hline
		MAE$\downarrow$ & 19.64$^\circ$ & 21.64$^\circ$ & 20.54$^\circ$ & 15.72$^\circ$ & 16.94$^\circ$ & \textcolor{red}{15.12}$^\circ$ \\
		\hline
	\end{tabular}
\end{table}

\begin{figure*}[!htbp]
	\centering
	\includegraphics[width=0.9\textwidth]{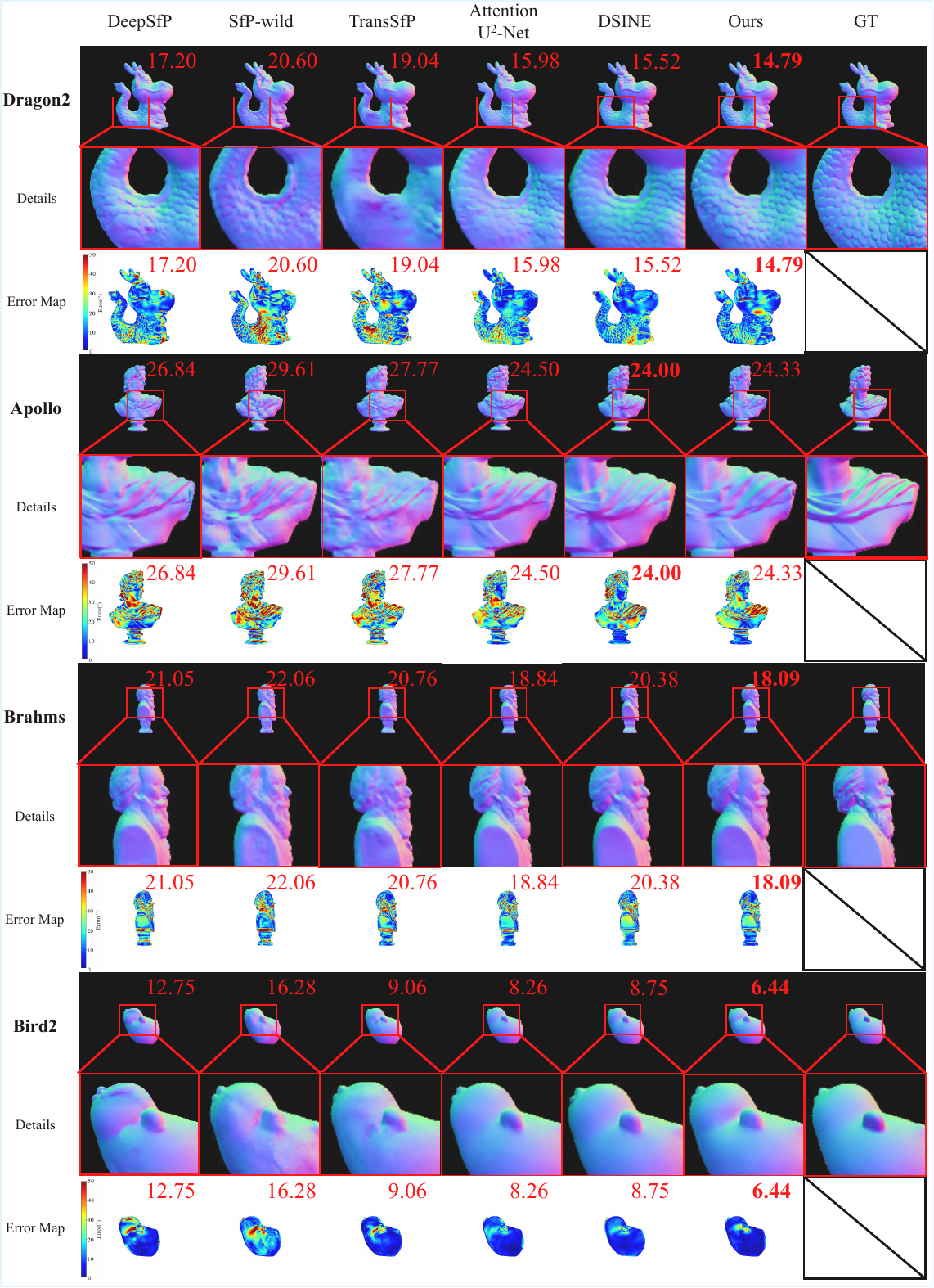}
	\caption{Comparison of different normal estimation methods on multiple samples under the same turbidity condition (10 ml emulsion). For each group, the results are presented in the order of predicted normal maps, enlarged local details, and angular error heatmaps. The red numbers indicate the MAE (°) of the corresponding method on each sample.}
	\label{fig:sample_comparison}
\end{figure*}

To further provide an intuitive comparison of 3D reconstruction performance under underwater scattering, Fig.~\ref{fig:sample_comparison} presents surface normal predictions on representative MuS-Polar3D test samples, together with zoomed-in local details and the corresponding error maps. Existing methods can generally recover the overall shape outline and basic geometry, but still suffer from noticeable over-smoothing or geometric distortion in regions with large curvature variation and rich high-frequency details (e.g., edges, folds, and local textures).

In contrast, the proposed method shows more stable geometric recovery across different objects and scattering levels. For detail-rich samples such as Dragon2 and Apollo, local surface undulations and continuity are better preserved, avoiding abrupt normal-direction changes. For portrait-like samples (Brahms) and small-scale targets (Bird2), the predictions also exhibit higher global consistency and finer detail depiction than the compared methods.

The error maps further support these observations. The errors of baseline methods mainly concentrate in regions with large curvature variations and often form noticeable structured patterns. In comparison, the proposed method yields lower overall error magnitudes, substantially fewer high-error regions, and a more uniform error distribution. This suggests that jointly modeling descattering and surface normal estimation within a unified framework can effectively mitigate systematic geometric inference errors induced by underwater scattering, thereby improving the stability and accuracy of normal prediction.

\begin{figure*}[!ht]
	\centering
	\includegraphics[width=\textwidth]{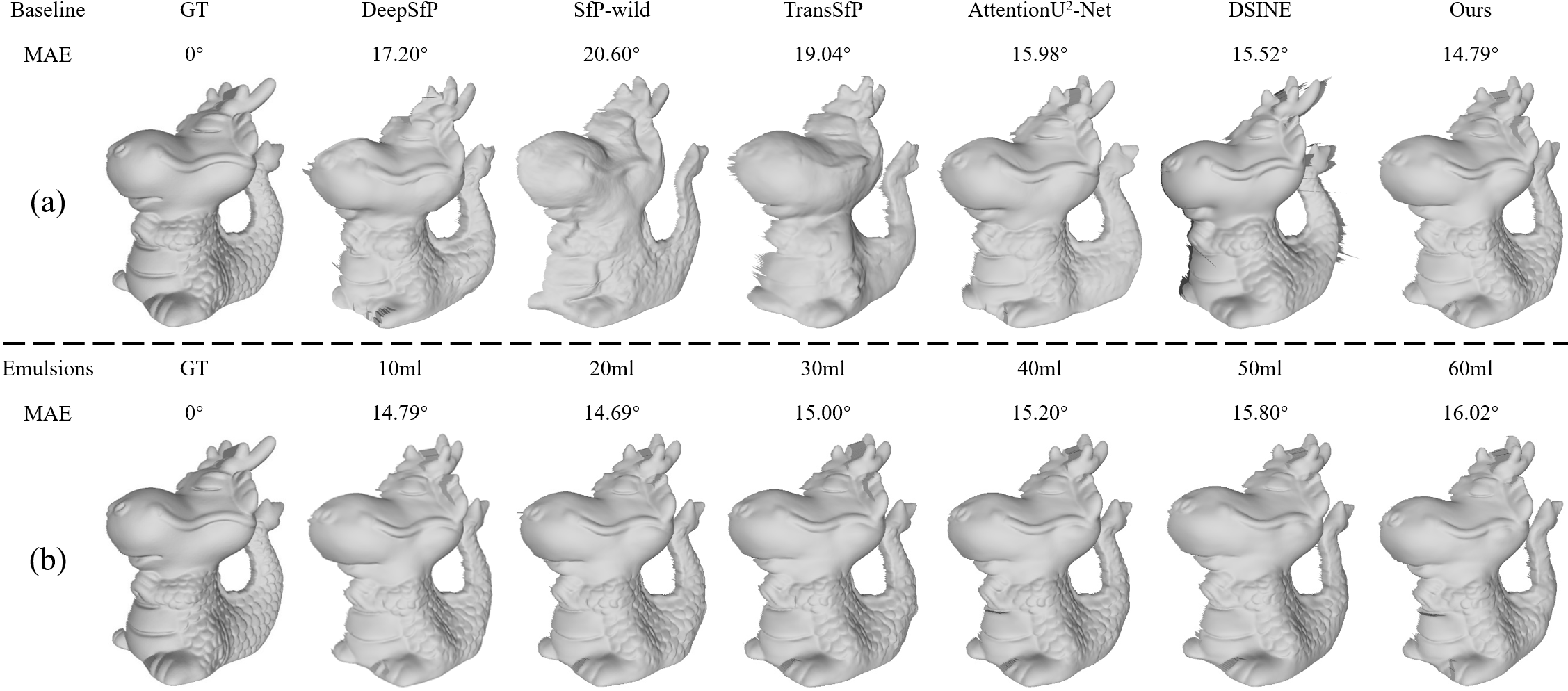}
	\caption{3D shape integration results from predicted normal maps using a normal integration method. (a) Integrated shapes of different baselines under 10\,ml emulsions. (b) Integrated shapes of the proposed method under different scattering conditions.}
	\label{fig:3D surface}
\end{figure*}

In addition, 3D shapes are recovered from the predicted normal maps using the method in~\cite{cao2022bilateral} to visualize the reconstruction quality. As shown in Fig.~\ref{fig:3D surface}, a smaller MAE generally corresponds to a more pronounced and complete integrated 3D shape. The proposed method produces more vivid geometry and richer texture details under different scattering conditions (Fig.~\ref{fig:3D surface}(b)), indicating strong robustness. However, the horn region in the example exhibits depth discontinuity in a single-view setting, which breaks integrability of the normal field and causes jumps in normal integration, leading to cliff-like reconstruction artifacts. Future work will address such depth discontinuities caused by boundary occlusions.

\subsection{Ablation study}

\begin{table}[h]
	\centering
	\caption{Ablation results of key sub-networks/modules. The metrics are mean/median normal angular error ($^\circ$); lower is better.}
	\label{tab:ablation}
	{\small
		\begin{tabular}{lcccccc}
			\hline
			Model & w/o PPN & w/o DN & \begin{tabular}{@{}c@{}}w/o PPN\\ \& DN\end{tabular} & w/o CE & w/o DEConv & Proposed \\
			\hline
			Mean & 16.72$^\circ$ & 15.37$^\circ$ & 15.56$^\circ$ & 15.46$^\circ$ & 23.03$^\circ$ & \textcolor{red}{15.12}$^\circ$ \\
			Median & 15.94$^\circ$ & 15.38$^\circ$ & 16.09$^\circ$ & 15.73$^\circ$ & 22.48$^\circ$ & \textcolor{red}{15.21}$^\circ$ \\
			\hline
		\end{tabular}
	}
\end{table}

To validate the effectiveness of key sub-networks and modules in the proposed structured learning framework, systematic ablation experiments are conducted on the MuS-Polar3D test set. Quantitative results are reported in Table~\ref{tab:ablation}. The experiments remove the polarization-parameter network (PPN), the descattering network (DN), both PPN and DN, as well as the color embedding module (CE) and detail-enhanced convolution (DEConv), and compare against the full model. This enables a quantitative analysis of how each component contributes to overall 3D reconstruction performance.

First, removing PPN increases the mean angular error from 15.12$^\circ$ to 16.72$^\circ$, indicating that predicting normals only from local descattered cues cannot fully exploit global geometric priors contained in polarization information. Removing DN yields a smaller but still noticeable degradation, suggesting that explicit low-level descattering modeling provides a more stable input for subsequent geometric inference under scattering. When both PPN and DN are removed, the model degenerates to end-to-end normal regression directly from scattered polarization images; the error remains higher than the full model, confirming the necessity of jointly modeling ``low-level restoration + high-level geometry''.

For architectural modules, removing CE leads to a clear increase in normal error, showing that the embedding mechanism based on the color--geometry isomorphism improves consistency and stability of normal estimation. The most severe drop occurs when removing DEConv: the mean angular error rises sharply to 23.03$^\circ$, indicating that explicit modeling of high-frequency geometric details and directional variations is critical for accurate underwater SfP task. Overall, these components complement each other within the unified framework and together yield the best performance.

\section{Conclusion}
Severe scattering in aquatic environments remains the fundamental challenge that cripples the performance of mainstream 3D optical imaging methods for underwater scenarios.While polarization imaging exhibits unparalleled dual advantages in both descattering and shape-from-polarization (SfP)-based 3D structure recovery, state-of-the-art methods are still limited by the sequential processing pipeline: descattering and 3D reconstruction are treated as two isolated tasks, leading to progressive error accumulation that cannot be corrected in subsequent stages. To address this fundamental limitation, this paper rethinks the underwater 3D imaging problem from the perspective of full-chain imaging pipeline optimization, and develops an end-to-end unified framework that enables joint learning and global optimization of polarization-guided descattering and 3D normal reconstruction. The proposed method effectively suppresses the error propagation in cascaded pipelines, and achieves significant performance improvements over conventional approaches for practical underwater scenarios.

UD-SfPNet is proposed to tightly couple polarization-based descattering with polarization-driven normal estimation. The framework consists of a polarization parameter network and a descattering network, which together provide inputs to the normal estimation network, enabling end-to-end joint modeling of low-level restoration and high-level geometric inference. Two key modules are further introduced to improve performance: a CE module, which exploits the mapping between color and geometry to enhance prediction consistency, and a DEConv module, which explicitly models high-frequency geometry and directional variations to better preserve subtle details in target regions. Guided by physical models, these modules, together with polarization features, are embedded into the network, allowing UD-SfPNet to produce both clearer images and more accurate surface normal predictions under underwater scattering.

Extensive experiments on the MuS-Polar3D dataset demonstrate the effectiveness of UD-SfPNet. The method achieves the lowest mean normal angular error (15.12$^\circ$) among all compared approaches, indicating a clear improvement in reconstruction accuracy. It also shows strong robustness and generalization across different scattering levels and target properties. Overall, by deeply integrating polarization physics with deep learning, this work establishes a new paradigm for underwater 3D imaging and provides a practical solution for underwater robotic vision and ocean-exploration imaging in challenging environments.

\section*{Declaration of generative AI and AI-assisted technologies in the manuscript preparation process}
During the preparation of this work the author(s) used ChatGPT (OpenAI) in order to improve the readability and language of the manuscript.
After using this tool/service, the author(s) reviewed and edited the content as needed and take(s) full responsibility for the content of the published article.

  \bibliographystyle{elsarticle-num} 
  \bibliography{reference.bib}



\end{document}